\documentclass[letterpaper]{article} 
\usepackage{aaai2026}  
\usepackage{times}  
\usepackage{helvet}  
\usepackage{courier}  
\usepackage[hyphens]{url}  
\usepackage{graphicx} 
\usepackage{natbib}  
\usepackage{caption} 
\usepackage{algorithm}
\usepackage{algpseudocode}
\usepackage{newfloat}
\usepackage{listings}
\DeclareCaptionStyle{ruled}{labelfont=normalfont,labelsep=colon,strut=off} 
\floatstyle{ruled}
\newfloat{listing}{tb}{lst}{}
\floatname{listing}{Listing}
\usepackage{amssymb}
\usepackage{amsmath}
\usepackage{booktabs}
\usepackage{multirow}
\usepackage[table,xcdraw]{xcolor}

\usepackage{xcolor}         

\title{\textsc{Prune\&Comp}: Free Lunch for Layer-Pruned LLMs via Iterative Pruning with Magnitude Compensation}
\author{
    Xinrui Chen\textsuperscript{\rm 1}, Hongxing Zhang\textsuperscript{\rm 2}, Fanyi Zeng\textsuperscript{\rm 1}, Yongxian Wei\textsuperscript{\rm 1}, 
    Yizhi Wang\textsuperscript{\rm 1}, \\ Xitong Ling\textsuperscript{\rm 1}, Guanghao Li\textsuperscript{\rm 1}, Chun Yuan\textsuperscript{\rm 1}\thanks{Corresponding author.}
}
\affiliations{
    \textsuperscript{\rm 1} Shenzhen International Graduate School, Tsinghua University \\
    \textsuperscript{\rm 2} School of Information Science and Technology, Guangdong University of Foreign Studies

    cxr22@tsinghua.org.cn, yuanc@sz.tsinghua.edu.cn
}

\usepackage{bibentry}

\begin{document}

\maketitle

\begin{abstract}
Layer pruning has emerged as a promising technique for compressing large language models (LLMs) while achieving acceleration proportional to the pruning ratio. In this work, we identify that removing any layer induces a significant magnitude gap in hidden states, resulting in substantial performance degradation. To address this issue, we propose \textsc{Prune\&Comp}, a novel plug-and-play layer pruning scheme that leverages magnitude compensation to mitigate such gaps in a training-free manner. Specifically, we first estimate the magnitude gap caused by layer removal and then eliminate this gap by rescaling the remaining weights offline, with zero runtime overhead incurred. We further demonstrate the advantages of \textsc{Prune\&Comp} through an iterative pruning strategy. When integrated with an iterative prune-and-compensate loop, \textsc{Prune\&Comp} consistently enhances existing layer pruning metrics. For instance, when 5 layers of LLaMA-3-8B are pruned using the prevalent block influence metric, \textsc{Prune\&Comp} nearly halves the perplexity and retains 93.19\% of the original model's question-answering performance, outperforming the baseline by 4.01\%.
\end{abstract}


\section{Introduction}


In recent years, large language models (LLMs) have achieved remarkable success across a wide range of natural-language-processing tasks~\cite{achiam2023gpt, jiang2023mistral, qwen3technicalreport, dubey2024llama3, team2025kimi, liu2024deepseek, guo2025deepseek}. As model size increases, LLMs exhibit substantial performance gains; however, the vast parameter count incurs prohibitive computational costs and long inference latency. The model compression community has thus introduced effective LLM compression schemes, primarily quantization, knowledge distillation, and pruning~\cite{sreenivas2024llm,muralidharan2024compact,ashkboos2024quarot, sun2024flatquant,ashkboos2024slicegpt, van2023llm, xia2023sheared, sarah2024llama, hu2024sp3}. Among these, pruning is an auspicious approach that reduces model size by removing unimportant parameters or components.

\begin{figure}[t]
\centering
\includegraphics[width=1.0\columnwidth]{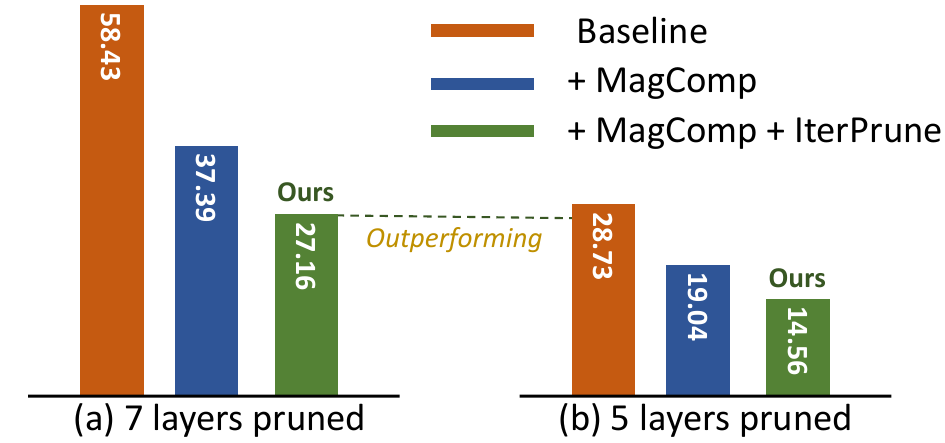} 
\caption{
Average perplexity (↓) of pruned LLaMA-3-8B on WikiText-2, C4, and PTB. Our 7-layer-pruned model outperforms the 5-layer-pruned baseline. Baseline: naive pruning using BI metric \cite{men2024shortgpt}; +MagComp: magnitude compensation; +IterPrune: iterative pruning.
}
\label{fig:motivation1}
\end{figure}

Structured pruning is the mainstream pruning paradigm. Unlike semi-structured sparsity or unstructured pruning, which irregularly eliminate weights or modules and therefore introduce irregular memory access, structured pruning avoids reliance on specialized hardware or software optimizations and delivers genuine acceleration. Within the structured pruning literature, the dominant techniques fall into depth pruning and width pruning. Width pruning removes unimportant weight channels and attention heads, thereby shrinking the width of LLMs. Depth pruning, also known as layer pruning, discards entire Transformer layers to reduce model depth. Existing studies~\cite{gromov2024unreasonable,kim2024shortened} show that, at moderate pruning ratios, layer pruning can retain most of the original performance without retraining. Moreover, by reducing model depth, layer pruning shortens LLM inference latency and achieves higher speed-ups at the same pruning ratio, without requiring hardware- or software-specific support.

The majority of layer pruning methods~\cite{gromov2024unreasonable,kim2024shortened,song2024sleb,chen2024compressing,men2024shortgpt} have designed sophisticated layer importance metrics to locate redundant layers. Representative works based on layer-output similarity~\cite{men2024shortgpt,chen2024compressing,gromov2024unreasonable}, performance-impact metrics~\cite{kim2024shortened,song2024sleb}, and gradient-based metrics~\cite{kim2024shortened,ma2023llm} have attracted considerable attention due to their strong empirical results. These methods typically identify and remove redundant layers in either a one-shot or iterative strategy. Iterative methods account for inter-layer dependencies and generally yield more competitive performance~\cite{kim2024shortened,song2024sleb}.

Nevertheless, even though existing layer pruning strategies preserve most performance without retraining, they still suffer from noticeable performance degradation. We investigate this phenomenon and offer two key observations:

\begin{figure*}[t]
\centering
\includegraphics[width=1\textwidth]{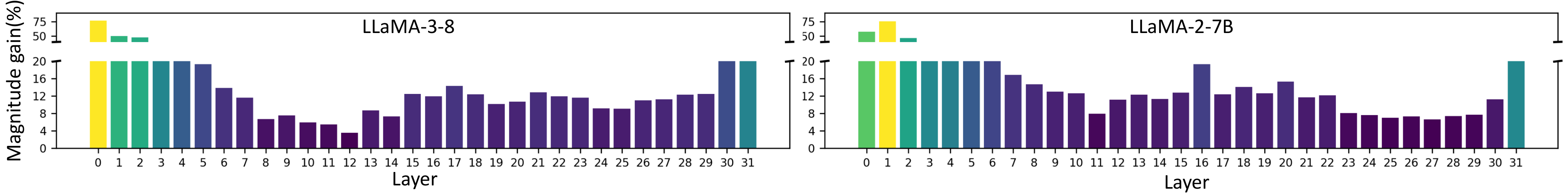} 
\caption{Visualization on the channel-wise averaged magnitude gain ratio of each layer. All layers produce magnitude gain.}
\label{fig:motivation2}
\end{figure*}

\begin{figure*}[t]
\centering
\includegraphics[width=1\textwidth]{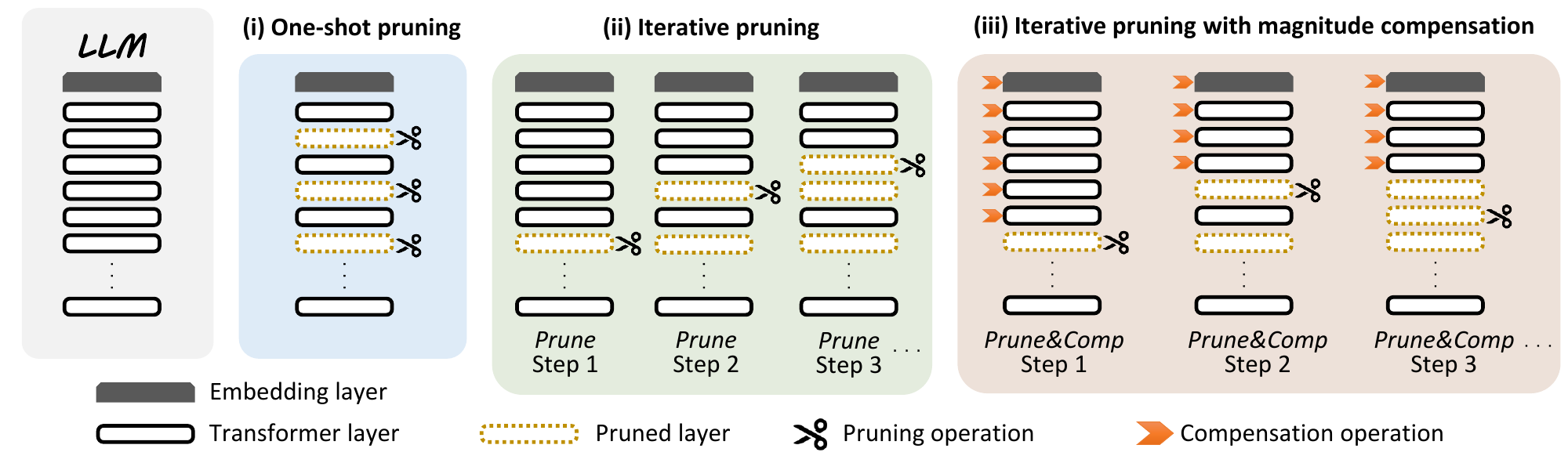} 
\caption{Comparison on conventional pruning strategy and the proposed \textsc{Prune\&Comp}.}
\label{fig:overview}
\end{figure*}

\noindent \textit{
(1) We demonstrate that removing any layer from an LLM introduces a significant gap in the magnitudes of hidden states, as shown in Figure \ref{fig:motivation2}}. This is an intrinsic property of LLMs and is independent of the pruning metric. Our intuition is to compensate for this magnitude gap, showing that a simple compensation strategy can easily restore model performance, as presented in Figure \ref{fig:motivation1}.

\noindent \textit{
(2) Although iterative layer pruning methods account for inter-layer dependencies, the already-damaged pruned model injects errors into subsequent layer importance assessments.} We show that a simple compensation operation during the iterative process allows the pruned model to better identify redundant layers and thus achieve superior results, as presented in Figure \ref{fig:motivation1}.

Building on these observations, we introduce a simple yet effective method, \textsc{Prune\&Comp}, which compensates for the magnitude gap caused by layer pruning via weight modifications that incur no online inference overhead. When combined with iterative layer pruning, \textsc{Prune\&Comp} improves the robustness of the pruned model and thus enhances the effectiveness of the iterative layer importance estimation. Our major contributions are as follows:

\begin{itemize}
\item We reveal that layer pruning inevitably introduces a large magnitude gap, causing severe performance drops in LLMs. This degradation is rooted in the intrinsic properties of the model, regardless of the pruning metric.
\item We introduce \textsc{Prune\&Comp}, a training-free recipe that couples iterative layer pruning with magnitude compensation to close the gap in a training-free manner.
\item Extensive experiments show that \textsc{Prune\&Comp} consistently boosts prevalent pruning metrics by a large margin, while adding no online inference overhead.
\end{itemize}

\section{Related Works}

\subsection{Width Pruning}

Width pruning for LLMs aims to reduce the computational footprint by selectively removing redundant or less informative channels, attention heads, and their coupled structures. LLM-Pruner~\cite{ma2023llm}, the first structured pruning framework tailored for LLMs, leverages gradient-based importance estimation to identify and remove non-essential coupled structures, achieving substantial compression while preserving core model capabilities.  
Sheared LLaMA~\cite{xia2023sheared} accelerates LLM pre-training by incorporating targeted structured pruning and dynamic batch loading, demonstrating improved training efficiency and outperforming baseline models of similar size.
Wanda~\cite{sun2023simple} introduces a retraining-free technique that induces sparsity by eliminating weights with the smallest product of magnitude and input hidden states, offering a simple yet effective approach to parameter reduction.  
Similarly, FLAP~\cite{an2024fluctuation} exploits activation fluctuations to a novel retraining-free structured pruning framework that enhances storage efficiency and inference speed.  

\subsection{Depth Pruning}

In contrast to width pruning, depth pruning, also known as layer pruning, compresses the LLMs by removing entire model layers, while retaining the parameter dimensions of the remaining components.
At the same pruning rate, layer pruning achieves higher speed-ups without additional architectural dependencies.
ShortGPT~\cite{men2024shortgpt} proposes Block Influence (BI), a layer-level importance metric that captures the semantic shift between a layer’s input and output representations.
LLM-Streamline~\cite{chen2024streamlining} leverages cosine similarity to assess layer importance and introduces a novel compression metric, stability, to quantify the structural sensitivity of models.  
SLEB~\cite{song2024sleb} exploits the redundancy across adjacent transformer blocks based on block-level Perplexity (PPL) evaluations.  
Shortened LLaMA~\cite{kim2024shortened} adopts a straightforward depth pruning approach, guided by gradient-based and magnitude-based metrics to determine removable layers.

\section{Motivation}

\subsection{Preliminaries on LLM Layer Pruning}
LLMs are predominantly built upon the Transformer architecture, which comprises a stack of Transformer decoder layers with residual connections. We denote the $\ell$-th Transformer layer as $f(X^{(\ell)},\theta^{(\ell)})$, where $X^{(\ell)}$ represents its input hidden states and $\theta^{(\ell)}$ signifies its corresponding parameters. Given the prevalent use of the pre-norm architecture in LLMs, the output of the $\ell$-th layer can be expressed as:
\begin{equation}
    X^{(\ell+1)}=X^{(\ell)}+f(X^{(\ell)},\theta^{(\ell)}).
\end{equation}

To remove LLM layers with index from $\ell^{*}$-th layer to $(\ell^{*}+n)$-th layer, 
we skip these layers and pass the input of the $\ell^{*}$-th layer to $(\ell^{*}+n)$-th layer, i.e., 
\begin{equation}
\label{eq:prune}
{X}^{(\ell^{*}+n)}={X}^{(\ell^{*})}+f({X}^{(\ell^{*})},\theta^{(\ell^{*}+n)}).
\end{equation}

Especially, $n = 1$ indicates the removal of the $\ell^*$-th layer.

\subsection{Layer Pruning Produces Magnitude Gap}

We quantitatively analyze the magnitude gain introduced by every individual LLM layer.
For each layer $\ell$, we calculate the channel-wise averaged magnitude gain ratio between its input and output hidden states over a calibration set. The magnitude gain ratio of layer $\ell$ is defined as:

\begin{equation}
\mathbf{\delta}^{(\ell)}=\left(\mathbb{E}_{({X}^{(\ell)},{X}^{(\ell+1)})\in \mathcal{D}}\frac{1}{C}\sum_k^C \frac{\|{X}_{:,k}^{(\ell+1)}\|_1}{\|{X}_{:,k}^{(\ell)}\|_1}-1\right)\times100\%,
\end{equation}

\noindent where $\mathcal{D}$ denotes the calibration set, and ${X}^{(\ell)}\in \mathbb{R}^{B \times T \times C}$ represents the input hidden state of the $\ell$-th layer with batch size $B$, length of tokens $T$ and hidden dimension $C$, obtained through calibration samples.
Figure \ref{fig:motivation2} visualizes the individual layer magnitude gain ratio of LLaMA-3-8B and LLaMA-2-7B. It is observed that each layer produces a pronounced increase in magnitude, and the largest surge exceeds 70\% and occurs in the early layers.
Consequently, removing any layer inevitably creates a corresponding magnitude gap, regardless of the pruning metrics. Our intuition is to compensate for this gap and preserve the scale of hidden states. As shown in Figure \ref{fig:motivation1}, simply compensating this magnitude gap (+MagComp) boosts the performance.

\subsection{Rethinking Iterative Layer Pruning}

Figure \ref{fig:overview} provides a detailed comparison diagram of one-shot pruning, iterative pruning, and the proposed \textsc{Prune\&Comp} scheme.
One-shot pruning makes its entire pruning decision in a single pass, offering high search efficiency. However, removing multiple layers simultaneously ignores inter-layer dependencies, and the cumulative impact can be far more destructive than the sum of the individual effects. Iterative pruning, by contrast, re-evaluates the layer importance after every removal. This procedure captures second-order interactions and generally yields a superior accuracy-compression trade-off~\cite{kim2024shortened,song2024sleb}. Nevertheless, each iterative pruning step damages the model, which in turn hampers robust assessment of the remaining layers. Motivated by this, we propose to compensate the pruned model during the iterative process, enabling it to locate redundant layers more precisely.

\begin{figure*}[t]
\centering
\includegraphics[width=0.97\textwidth]{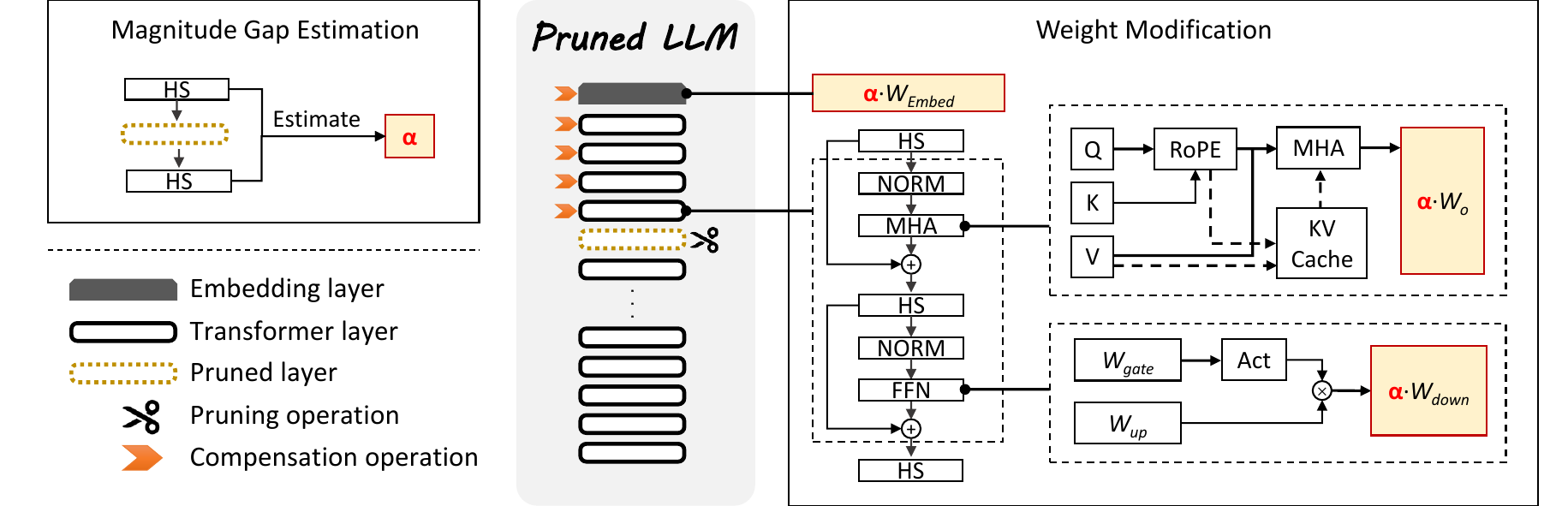} 
\caption{Magnitude compensation applied to LLM. {HS}: hidden states; {MHA}: multi-head attention; {FFN}: feed-forward network;  $W_{Embed}$: embedding weight; NORM: normalization layer; {RoPE}: rotary position embedding; ${W_{gate}}, {W_{up}}, {W_{down}}$: FFN gate, up and down projection weight, respectively; ${W_O}$: MHA output projection weight; {Act}: activation function.}
\label{fig:magcomp}
\end{figure*}

\section{Methods}

\subsection{Layer Pruning Metrics}
\label{sec:metric}

In this work, the following five prevalent layer pruning metrics are employed to identify redundant layers in LLMs.

\paragraph{Cosine Similarity-based Metrics.}
\begin{itemize}
\item \textbf{CosSim(BI)} \cite{men2024shortgpt}: Block Influence (BI) measures the importance of each layer by evaluating the similarity between a layer's input and its output. The BI score for the $\ell$-th layer can be calculated as follows:
\begin{equation}
BI_{\ell}={\mathbb{E}}_{X,t}\frac{(X^{(\ell)}_{t})^{T}X^{(\ell+1)}_{t}}{||X^{(\ell)}_{t}||_{2}||X^{(\ell+1)}_{t}||_{2}},
\end{equation}
where $X^{(\ell)}_{t}$ denotes the hidden states from the $t$-th token position of the $\ell$-th layer's input hidden states.
A higher BI score indicates a high cosine similarity between $X^{(\ell)}$ and $X^{(\ell+1)}$, suggesting redundancy of layer $\ell$.


\item \textbf{CosSim(CL)} \cite{chen2024compressing,gromov2024unreasonable}: Similarly, a series of contiguous layers (CL) with high cosine similarity between their input and output indicates redundancy.
When pruning $n$ layers, the importance of $n$ consecutive layers from layer $\ell$ to layer $\ell+n$ can be calculated as:
\begin{equation}
CL_{\ell,n}=\mathbb{E}_{X,t}\frac{(X^{(\ell)}_{t})^{T}X^{(\ell+n)}_{t}}{||X^{(\ell)}_{t}||_{2}||X^{(\ell+n)}_{t}||_{2}},
\end{equation}

\end{itemize}

\paragraph{Perplexity (PPL)} \cite{song2024sleb,kim2024shortened}: Lower PPL indicates the fluency of LLM text generation. Redundant blocks contribute less to the model performance, and their removal leads to a smaller increase in PPL. 
The PPL importance score $I_{PPL}^{n}$ of the $n$-th block is defined as:
\begin{equation}
    I_{PPL}^{\ell}= \text{exp}\left\{-\frac{1}{ST}\sum_{s=1}^{S}\sum_{t=1}^{T}\log~p_{\theta^{\ell}}(x_{t}^{(s)}|x_{<t}^{(s)})\right\},
\end{equation}
where $\theta^{\ell}$ denotes the model without its $\ell$-th block, $s=1,\dots,S$ are the indices for sequences, and $t=1,\dots,T$ are the indices for tokens in calibration set $D$. 

\paragraph{Taylor+} \cite{kim2024shortened}: Taylor metrics estimate the significance of a weight parameter by assessing the error caused by its removal. For a given calibration dataset $D$, this can be expressed as the alteration in the training loss $\mathcal{L}$:
$$|\mathcal{L}(W_{i,j}^{k,\ell};D)-\mathcal{L}(W_{i,j}^{k,\ell}=0;D)|\approx \left|\frac{\partial\mathcal{L}(D)}{\partial W_{i,j}^{k,\ell}}W_{i,j}^{k,\ell}\right|,$$
where second-order derivatives are omitted. The Taylor+ block importance score $I_{Taylor}^{\ell}$ can be defined as:
\begin{equation}
    I_{Taylor}^{\ell}= \sum_{k}\sum_{i}\sum_{j}\left|\frac{\partial\mathcal{L}(D)}{\partial W_{i,j}^{k,\ell}}W_{i,j}^{k,\ell}\right|,
\end{equation}
where $W^{k,\ell}$ is the linear weight matrix of operation type $k$ in the $\ell$-th Transformer block, and $W_{i,j}^{k,\ell}$ is its element. The Taylor+ method builds upon the Taylor metric by preserving the first four and the last two Transformer blocks, as their removal leads to severe performance drops. A lower $I_{Taylor}^{\ell}$ score indicates a block that is less important and therefore more suitable for pruning.

\paragraph{Magnitude+ (Mag+)} \cite{kim2024shortened}: Mag+ incorporates a heuristic rule based on the Magnitude metric (referencing Li et al., 2017b, which assumes weights with smaller norms are less informative) and preserves the first four and the last two blocks of the model. The underlying Mag+ importance score can be calculated as:
\begin{equation}
I_{Magnitude}^{\ell}=\sum_{k}\sum_{i}\sum_{j}|W_{i,j}^{k,\ell}|,
\end{equation}
where $W_{i,j}^{k,\ell}$ is an element of the linear weight matrix of operation type $k$ in the $\ell$-th Transformer block. A lower $I_{Magnitude}^{\ell}$ score indicates a block that is less important and therefore more suitable for pruning.

\subsection{Magnitude Compensation}

\begin{table*}[!th]
\centering
\caption{Performance comparison on perplexity (PPL) benchmark. Sparsity is indicated by layers pruned/total layers.}
\label{tab:table1-ppl}
\resizebox{\textwidth}{!}{%
\begin{tabular}{@{}clcccc|clcccc@{}}
\toprule
\multicolumn{6}{c|}{LLaMA-2-7B}                                                                                                                                                                                                     & \multicolumn{6}{c}{LLaMA-3-8B}                                                                                                                                                                                                        \\ \midrule
\textbf{Sparsity}       & \multicolumn{1}{c}{\textbf{Metric}}  & \multicolumn{1}{c}{\textbf{WikiText-2}} & \multicolumn{1}{c}{\textbf{C4}}        & \multicolumn{1}{c}{\textbf{PTB}}       & \multicolumn{1}{c|}{\textbf{Average}}  & \textbf{Sparsity}       & \multicolumn{1}{c}{\textbf{Metric}}  & \multicolumn{1}{c}{\textbf{WikiText-2}} & \multicolumn{1}{c}{\textbf{C4}}        & \multicolumn{1}{c}{\textbf{PTB}}        & \multicolumn{1}{c}{\textbf{Average}}    \\ \midrule
-                       & Dense                                & 5.47                                    & 6.97                                   & 22.51                                  & 11.65                                  & -                       & Dense                                & 6.14                                    & 8.88                                   & 10.59                                   & 8.54                                    \\ \midrule
                        & PPL                                  & 9.81                                    & 12.36                                  & 48.53                                  & 23.57                                  &                         & PPL                                  & 12.37                                   & 15.28                                  & 18.91                                   & 15.52                                   \\
                        & \cellcolor[HTML]{FEF2F0}+\textsc{Prune\&Comp} & \cellcolor[HTML]{FEF2F0}\textbf{8.57}   & \cellcolor[HTML]{FEF2F0}\textbf{10.55} & \cellcolor[HTML]{FEF2F0}\textbf{40.51} & \cellcolor[HTML]{FEF2F0}\textbf{19.88} &                         & \cellcolor[HTML]{FEF2F0}+\textsc{Prune\&Comp} & \cellcolor[HTML]{FEF2F0}\textbf{9.65}   & \cellcolor[HTML]{FEF2F0}\textbf{13.20} & \cellcolor[HTML]{FEF2F0}\textbf{16.02}  & \cellcolor[HTML]{FEF2F0}\textbf{12.96}  \\
                        & CosSim(CL)                           & 18.45                                   & 20.99                                  & 62.18                                  & 33.87                                  &                         & CosSim(CL)                           & 21.14                                   & 24.13                                  & 37.41                                   & 27.56                                   \\
                        & \cellcolor[HTML]{FEF2F0}+\textsc{Prune\&Comp} & \cellcolor[HTML]{FEF2F0}\textbf{13.78}  & \cellcolor[HTML]{FEF2F0}\textbf{15.31} & \cellcolor[HTML]{FEF2F0}\textbf{49.40} & \cellcolor[HTML]{FEF2F0}\textbf{26.16} &                         & \cellcolor[HTML]{FEF2F0}+\textsc{Prune\&Comp} & \cellcolor[HTML]{FEF2F0}\textbf{16.90}  & \cellcolor[HTML]{FEF2F0}\textbf{18.57} & \cellcolor[HTML]{FEF2F0}\textbf{21.64}  & \cellcolor[HTML]{FEF2F0}\textbf{19.04}  \\
                        & Mag+                                 & 49.39                                   & 34.65                                  & 184.78                                 & 89.61                                  &                         & Mag+                                 & 37.57                                   & 34.99                                  & 60.80                                   & 44.45                                   \\
                        & \cellcolor[HTML]{FEF2F0}+\textsc{Prune\&Comp} & \cellcolor[HTML]{FEF2F0}\textbf{11.73}  & \cellcolor[HTML]{FEF2F0}\textbf{13.28} & \cellcolor[HTML]{FEF2F0}\textbf{59.60} & \cellcolor[HTML]{FEF2F0}\textbf{28.20} &                         & \cellcolor[HTML]{FEF2F0}+\textsc{Prune\&Comp} & \cellcolor[HTML]{FEF2F0}\textbf{13.60}  & \cellcolor[HTML]{FEF2F0}\textbf{18.24} & \cellcolor[HTML]{FEF2F0}\textbf{22.52}  & \cellcolor[HTML]{FEF2F0}\textbf{18.12}  \\
                        & Taylor+                              & 18.45                                   & 20.99                                  & 63.02                                  & 34.15                                  &                         & Taylor+                              & 602.96                                  & 388.41                                 & 546.98                                  & 512.78                                  \\
                        & \cellcolor[HTML]{FEF2F0}+\textsc{Prune\&Comp} & \cellcolor[HTML]{FEF2F0}\textbf{10.61}  & \cellcolor[HTML]{FEF2F0}\textbf{12.10} & \cellcolor[HTML]{FEF2F0}\textbf{51.02} & \cellcolor[HTML]{FEF2F0}\textbf{24.58} &                         & \cellcolor[HTML]{FEF2F0}+\textsc{Prune\&Comp} & \cellcolor[HTML]{FEF2F0}\textbf{12.78}  & \cellcolor[HTML]{FEF2F0}\textbf{16.20} & \cellcolor[HTML]{FEF2F0}\textbf{20.03}  & \cellcolor[HTML]{FEF2F0}\textbf{16.34}  \\
                        & CosSim(BI)                           & 18.45                                   & 20.99                                  & 62.18                                  & 33.87                                  &                         & CosSim(BI)                           & 27.33                                   & 27.06                                  & 31.81                                   & 28.73                                   \\
\multirow{-10}{*}{7/32} & \cellcolor[HTML]{FEF2F0}+\textsc{Prune\&Comp} & \cellcolor[HTML]{FEF2F0}\textbf{11.45}  & \cellcolor[HTML]{FEF2F0}\textbf{12.96} & \cellcolor[HTML]{FEF2F0}\textbf{42.29} & \cellcolor[HTML]{FEF2F0}\textbf{22.23} & \multirow{-10}{*}{5/32} & \cellcolor[HTML]{FEF2F0}+\textsc{Prune\&Comp} & \cellcolor[HTML]{FEF2F0}\textbf{11.87}  & \cellcolor[HTML]{FEF2F0}\textbf{14.87} & \cellcolor[HTML]{FEF2F0}\textbf{16.93}  & \cellcolor[HTML]{FEF2F0}\textbf{14.56}  \\ \midrule
                        & PPL                                  & 14.91                                   & 17.03                                  & 67.73                                  & 33.22                                  &                         & PPL                                  & 15.08                                   & 17.57                                  & 22.09                                   & 18.2                                    \\
                        & \cellcolor[HTML]{FEF2F0}+\textsc{Prune\&Comp} & \cellcolor[HTML]{FEF2F0}\textbf{10.23}  & \cellcolor[HTML]{FEF2F0}\textbf{12.08} & \cellcolor[HTML]{FEF2F0}\textbf{50.96} & \cellcolor[HTML]{FEF2F0}\textbf{24.42} &                         & \cellcolor[HTML]{FEF2F0}+\textsc{Prune\&Comp} & \cellcolor[HTML]{FEF2F0}\textbf{12.40}  & \cellcolor[HTML]{FEF2F0}\textbf{15.98} & \cellcolor[HTML]{FEF2F0}\textbf{20.27}  & \cellcolor[HTML]{FEF2F0}\textbf{16.22}  \\
                        & CosSim(CL)                           & 35.68                                   & 36.10                                  & 96.52                                  & 56.10                                  &                         & CosSim(CL)                           & 2287.73                                 & 1491.37                                & 4738.81                                 & 2839.30                                 \\
                        & \cellcolor[HTML]{FEF2F0}+\textsc{Prune\&Comp} & \cellcolor[HTML]{FEF2F0}\textbf{19.37}  & \cellcolor[HTML]{FEF2F0}\textbf{20.13} & \cellcolor[HTML]{FEF2F0}\textbf{58.20} & \cellcolor[HTML]{FEF2F0}\textbf{32.57} &                         & \cellcolor[HTML]{FEF2F0}+\textsc{Prune\&Comp} & \cellcolor[HTML]{FEF2F0}\textbf{204.06} & \cellcolor[HTML]{FEF2F0}\textbf{231.6} & \cellcolor[HTML]{FEF2F0}\textbf{256.53} & \cellcolor[HTML]{FEF2F0}\textbf{230.73} \\
                        & Mag+                                 & 362.15                                  & 48.79                                  & 273.07                                 & 228.00                                 &                         & Mag+                                 & 40.70                                   & 36.95                                  & 44.85                                   & 40.83                                   \\
                        & \cellcolor[HTML]{FEF2F0}+\textsc{Prune\&Comp} & \cellcolor[HTML]{FEF2F0}\textbf{19.09}  & \cellcolor[HTML]{FEF2F0}\textbf{18.88} & \cellcolor[HTML]{FEF2F0}\textbf{95.98} & \cellcolor[HTML]{FEF2F0}\textbf{44.65} &                         & \cellcolor[HTML]{FEF2F0}+\textsc{Prune\&Comp} & \cellcolor[HTML]{FEF2F0}\textbf{33.33}  & \cellcolor[HTML]{FEF2F0}\textbf{35.02} & \cellcolor[HTML]{FEF2F0}\textbf{42.93}  & \cellcolor[HTML]{FEF2F0}\textbf{37.09}  \\
                        & Taylor+                              & 35.68                                   & 36.10                                  & 96.52                                  & 56.10                                  &                         & Taylor+                              & 2287.86                                 & 1491.38                                & 4741.9                                  & 2840.38                                 \\
                        & \cellcolor[HTML]{FEF2F0}+\textsc{Prune\&Comp} & \cellcolor[HTML]{FEF2F0}\textbf{13.80}  & \cellcolor[HTML]{FEF2F0}\textbf{14.52} & \cellcolor[HTML]{FEF2F0}\textbf{69.04} & \cellcolor[HTML]{FEF2F0}\textbf{32.45} &                         & \cellcolor[HTML]{FEF2F0}+\textsc{Prune\&Comp} & \cellcolor[HTML]{FEF2F0}\textbf{21.10}  & \cellcolor[HTML]{FEF2F0}\textbf{22.05} & \cellcolor[HTML]{FEF2F0}\textbf{32.47}  & \cellcolor[HTML]{FEF2F0}\textbf{25.21}  \\
                        & CosSim(BI)                           & 35.68                                   & 36.10                                  & 96.52                                  & 56.10                                  &                         & CosSim(BI)                           & 57.76                                   & 50.13                                  & 67.39                                   & 58.43                                   \\
\multirow{-10}{*}{9/32} & \cellcolor[HTML]{FEF2F0}+\textsc{Prune\&Comp} & \cellcolor[HTML]{FEF2F0}\textbf{18.53}  & \cellcolor[HTML]{FEF2F0}\textbf{17.99} & \cellcolor[HTML]{FEF2F0}\textbf{60.38} & \cellcolor[HTML]{FEF2F0}\textbf{32.30} & \multirow{-10}{*}{7/32} & \cellcolor[HTML]{FEF2F0}+\textsc{Prune\&Comp} & \cellcolor[HTML]{FEF2F0}\textbf{28.43}  & \cellcolor[HTML]{FEF2F0}\textbf{24.48} & \cellcolor[HTML]{FEF2F0}\textbf{28.57}  & \cellcolor[HTML]{FEF2F0}\textbf{27.16}  \\ \bottomrule
\end{tabular}%
}
\end{table*}
\begin{table*}[!ht]
\centering
\caption{Performance comparison on question answering (QA) benchmark. Sparsity is indicated by layers pruned/total layers (compression rate). RP denotes the  relative performance (\%).}
\label{tab:table2-qa}
\resizebox{2\columnwidth}{!}{%
\begin{tabular}{@{}cclccccccccccc@{}}
\toprule
\textbf{Model}                         & \textbf{Sparsity}                          & \multicolumn{1}{c}{\textbf{Metric}}  & \multicolumn{1}{c}{\textbf{ARC-c}}     & \multicolumn{1}{c}{\textbf{ARC-e}}     & \multicolumn{1}{c}{\textbf{BoolQ}}     & \textbf{CoPa}                          & \textbf{HeSw}                          & \textbf{PIQA}                          & \textbf{Race-h}                        & \textbf{WG}                            & \textbf{WSC}                           & \textbf{Average}                       & \textbf{RP}                            \\ \cmidrule(r){1-14}
                                       & \textbf{0/32}                              & Dense                                & 53.41                                  & 77.78                                  & 81.28                                  & 89.00                                  & 79.16                                  & 80.85                                  & 40.19                                  & 72.85                                  & 86.45                                  & 73.44                                  & 100.00                                 \\ \cmidrule(lr){2-14}
                                       &                                            & PPL                                  & 32.76                                  & 61.36                                  & \textbf{56.42}                         & 75.00                                  & 61.77                                  & 75.52                                  & 32.25                                  & 54.22                                  & 65.20                                  & 57.17                                  & 77.84                                  \\
                                       &                                            & \cellcolor[HTML]{FEF2F0}+\textsc{Prune\&Comp} & \cellcolor[HTML]{FEF2F0}\textbf{40.87} & \cellcolor[HTML]{FEF2F0}\textbf{66.58} & \cellcolor[HTML]{FEF2F0}56.27 & \cellcolor[HTML]{FEF2F0}\textbf{85.00} & \cellcolor[HTML]{FEF2F0}\textbf{67.44} & \cellcolor[HTML]{FEF2F0}\textbf{76.22} & \cellcolor[HTML]{FEF2F0}\textbf{33.97} & \cellcolor[HTML]{FEF2F0}\textbf{62.59} & \cellcolor[HTML]{FEF2F0}\textbf{73.26} & \cellcolor[HTML]{FEF2F0}\textbf{62.47} & \cellcolor[HTML]{FEF2F0}\textbf{85.06} \\
                                       &                                            & CosSim(CL)                           & 47.35                                  & 66.20                                  & 73.52                                  & 84.00                                  & 71.10                                  & 74.27                                  & 36.65                                  & 71.03                                  & 76.56                                  & 66.74                                  & 90.88                                  \\
                                       &                                            & \cellcolor[HTML]{FEF2F0}+\textsc{Prune\&Comp} & \cellcolor[HTML]{FEF2F0}\textbf{48.63} & \cellcolor[HTML]{FEF2F0}\textbf{69.99} & \cellcolor[HTML]{FEF2F0}\textbf{74.07} & \cellcolor[HTML]{FEF2F0}\textbf{85.00} & \cellcolor[HTML]{FEF2F0}\textbf{72.63} & \cellcolor[HTML]{FEF2F0}\textbf{75.95} & \cellcolor[HTML]{FEF2F0}\textbf{37.51} & \cellcolor[HTML]{FEF2F0}\textbf{72.61} & \cellcolor[HTML]{FEF2F0}\textbf{79.12} & \cellcolor[HTML]{FEF2F0}\textbf{68.39} & \cellcolor[HTML]{FEF2F0}\textbf{93.12} \\
                                       &                                            & Mag+                                 & 29.95                                  & 56.36                                  & 53.21                                  & 73.00                                  & 40.35                                  & 69.64                                  & 27.18                                  & 52.80                                  & 60.44                                  & 51.44                                  & 70.04                                  \\
                                       &                                            & \cellcolor[HTML]{FEF2F0}+\textsc{Prune\&Comp} & \cellcolor[HTML]{FEF2F0}\textbf{32.76} & \cellcolor[HTML]{FEF2F0}\textbf{57.87} & \cellcolor[HTML]{FEF2F0}\textbf{58.96} & \cellcolor[HTML]{FEF2F0}\textbf{82.00} & \cellcolor[HTML]{FEF2F0}\textbf{59.85} & \cellcolor[HTML]{FEF2F0}\textbf{73.23} & \cellcolor[HTML]{FEF2F0}\textbf{30.72} & \cellcolor[HTML]{FEF2F0}\textbf{55.25} & \cellcolor[HTML]{FEF2F0}\textbf{66.30} & \cellcolor[HTML]{FEF2F0}\textbf{57.44} & \cellcolor[HTML]{FEF2F0}\textbf{78.21} \\
                                       &                                            & Taylor+                              & 33.53                                  & 45.58                                  & 55.00                         & 55.00                         & 35.86                                  & 59.52                                  & 24.31                                  & 60.85                                  & 65.57                                  & 48.36                                  & 65.85                                  \\
                                       &                                            & \cellcolor[HTML]{FEF2F0}+\textsc{Prune\&Comp} & \cellcolor[HTML]{FEF2F0}\textbf{46.16} & \cellcolor[HTML]{FEF2F0}\textbf{70.24} & \cellcolor[HTML]{FEF2F0}\textbf{69.36} & \cellcolor[HTML]{FEF2F0}\textbf{81.00} & \cellcolor[HTML]{FEF2F0}\textbf{69.67} & \cellcolor[HTML]{FEF2F0}\textbf{73.83} & \cellcolor[HTML]{FEF2F0}\textbf{37.99} & \cellcolor[HTML]{FEF2F0}\textbf{70.56} & \cellcolor[HTML]{FEF2F0}\textbf{79.85} & \cellcolor[HTML]{FEF2F0}\textbf{66.52} & \cellcolor[HTML]{FEF2F0}\textbf{90.57} \\
                                       &                                            & CosSim(BI)                           & 45.56                                  & 63.51                                  & \textbf{73.12}                         & 79.00                                  & 70.13                                  & 74.92                                  & 36.94                                  & 71.19                                  & 75.09                                  & 65.50                                  & 89.18                                  \\
                                       & \multirow{-10}{*}{\textbf{5/32 (13.58\%)}} & \cellcolor[HTML]{FEF2F0}+\textsc{Prune\&Comp} & \cellcolor[HTML]{FEF2F0}\textbf{46.50} & \cellcolor[HTML]{FEF2F0}\textbf{70.54} & \cellcolor[HTML]{FEF2F0}71.31 & \cellcolor[HTML]{FEF2F0}\textbf{84.00} & \cellcolor[HTML]{FEF2F0}\textbf{72.43} & \cellcolor[HTML]{FEF2F0}\textbf{76.01} & \cellcolor[HTML]{FEF2F0}\textbf{37.99} & \cellcolor[HTML]{FEF2F0}\textbf{73.32} & \cellcolor[HTML]{FEF2F0}\textbf{83.88} & \cellcolor[HTML]{FEF2F0}\textbf{68.44} & \cellcolor[HTML]{FEF2F0}\textbf{93.19} \\ \cmidrule(lr){2-14}
                                       &                                            & PPL                                  & 32.76                                  & 58.84                                  & 45.38                                  & 75.00                                  & \textbf{59.22}                         & \textbf{73.56}                         & \textbf{30.72}                         & 53.83                                  & \textbf{67.77}                         & 55.23                                  & 75.20                                  \\
                                       &                                            & \cellcolor[HTML]{FEF2F0}+\textsc{Prune\&Comp} & \cellcolor[HTML]{FEF2F0}\textbf{33.53} & \cellcolor[HTML]{FEF2F0}\textbf{60.48} & \cellcolor[HTML]{FEF2F0}\textbf{47.52} & \cellcolor[HTML]{FEF2F0}\textbf{76.00} & \cellcolor[HTML]{FEF2F0}59.03          & \cellcolor[HTML]{FEF2F0}\textbf{73.56} & \cellcolor[HTML]{FEF2F0}\textbf{30.72} & \cellcolor[HTML]{FEF2F0}\textbf{54.54} & \cellcolor[HTML]{FEF2F0}67.40          & \cellcolor[HTML]{FEF2F0}\textbf{55.86} & \cellcolor[HTML]{FEF2F0}\textbf{76.07} \\
                                       &                                            & CosSim(CL)                           & 28.92                                  & 39.56                                  & 38.07                                  & 60.00                                  & 33.26                                  & 59.47                                  & 24.02                                  & 55.56                                  & 59.71                                  & 44.29                                  & 60.30                                  \\
                                       &                                            & \cellcolor[HTML]{FEF2F0}+\textsc{Prune\&Comp} & \cellcolor[HTML]{FEF2F0}\textbf{32.76} & \cellcolor[HTML]{FEF2F0}\textbf{45.62} & \cellcolor[HTML]{FEF2F0}\textbf{52.20} & \cellcolor[HTML]{FEF2F0}\textbf{66.00} & \cellcolor[HTML]{FEF2F0}\textbf{43.41} & \cellcolor[HTML]{FEF2F0}\textbf{63.55} & \cellcolor[HTML]{FEF2F0}\textbf{27.66} & \cellcolor[HTML]{FEF2F0}\textbf{57.85} & \cellcolor[HTML]{FEF2F0}\textbf{62.64} & \cellcolor[HTML]{FEF2F0}\textbf{50.19} & \cellcolor[HTML]{FEF2F0}\textbf{68.34} \\
                                       &                                            & Mag+                                 & 25.60                                  & 46.04                                  & 56.18                                  & 70.00                                  & 43.36                                  & 64.91                                  & 27.46                                  & 53.43                                  & 55.31                                  & 49.14                                  & 66.92                                  \\
                                       &                                            & \cellcolor[HTML]{FEF2F0}+\textsc{Prune\&Comp} & \cellcolor[HTML]{FEF2F0}\textbf{30.63} & \cellcolor[HTML]{FEF2F0}\textbf{49.37} & \cellcolor[HTML]{FEF2F0}\textbf{58.10} & \cellcolor[HTML]{FEF2F0}\textbf{76.00} & \cellcolor[HTML]{FEF2F0}\textbf{51.31} & \cellcolor[HTML]{FEF2F0}\textbf{68.82} & \cellcolor[HTML]{FEF2F0}\textbf{28.71} & \cellcolor[HTML]{FEF2F0}\textbf{55.01} & \cellcolor[HTML]{FEF2F0}\textbf{60.81} & \cellcolor[HTML]{FEF2F0}\textbf{53.20} & \cellcolor[HTML]{FEF2F0}\textbf{72.43} \\
                                       &                                            & Taylor+                              & 29.01                                  & 39.56                                  & 38.00                          & 60.00                                  & 33.24                                  & 59.30                                  & 24.02                                  & 55.49                                  & 59.71                                  & 44.26                                  & 60.27                                  \\
                                       &                                            & \cellcolor[HTML]{FEF2F0}+\textsc{Prune\&Comp} & \cellcolor[HTML]{FEF2F0}\textbf{42.66} & \cellcolor[HTML]{FEF2F0}\textbf{67.51} & \cellcolor[HTML]{FEF2F0}\textbf{67.61} & \cellcolor[HTML]{FEF2F0}\textbf{78.00} & \cellcolor[HTML]{FEF2F0}\textbf{65.84} & \cellcolor[HTML]{FEF2F0}\textbf{72.36} & \cellcolor[HTML]{FEF2F0}\textbf{37.22} & \cellcolor[HTML]{FEF2F0}\textbf{70.01} & \cellcolor[HTML]{FEF2F0}\textbf{77.66} & \cellcolor[HTML]{FEF2F0}\textbf{64.32} & \cellcolor[HTML]{FEF2F0}\textbf{87.58} \\
                                       &                                            & CosSim(BI)                           & \textbf{42.41}                         & 56.65                                  & 65.26                                  & 75.00                                  & 64.70                                  & 70.89                                  & 34.16                                  & \textbf{71.19}                         & 73.63                                  & 61.54                                  & 83.80                                  \\
\multirow{-21}{*}{\rotatebox[origin=c]{90}{LLaMA-3-8B}} & \multirow{-10}{*}{\textbf{7/32 (19.01\%)}} & \cellcolor[HTML]{FEF2F0}+\textsc{Prune\&Comp} & \cellcolor[HTML]{FEF2F0}41.55 & \cellcolor[HTML]{FEF2F0}\textbf{62.37} & \cellcolor[HTML]{FEF2F0}\textbf{72.78} & \cellcolor[HTML]{FEF2F0}\textbf{80.00} & \cellcolor[HTML]{FEF2F0}\textbf{65.44} & \cellcolor[HTML]{FEF2F0}\textbf{71.38} & \cellcolor[HTML]{FEF2F0}\textbf{36.65} & \cellcolor[HTML]{FEF2F0}71.11          & \cellcolor[HTML]{FEF2F0}\textbf{78.02} & \cellcolor[HTML]{FEF2F0}\textbf{64.37} & \cellcolor[HTML]{FEF2F0}\textbf{87.64} \\ \cmidrule(r){1-14}
                                       & \textbf{0/32}                              & Dense                                & 56.57                                  & 80.93                                  & 86.57                                  & 85.00                                  & 74.93                                  & 77.80                                  & 40.96                                  & 67.80                                  & 83.15                                  & 72.63                                  & 100.00                                 \\ \cmidrule{2-14}
                                       &                                            & PPL                                  & \textbf{46.93}                         & \textbf{74.41}                         & 69.63                                  & \textbf{82.00}                         & \textbf{76.06}                         & \textbf{77.20}                         & \textbf{38.37}                         & \textbf{61.25}                         & \textbf{75.46}                         & \textbf{66.81}                         & \textbf{91.98}                         \\
                                       &                                            & \cellcolor[HTML]{FEF2F0}+\textsc{Prune\&Comp} & \cellcolor[HTML]{FEF2F0}46.84 & \cellcolor[HTML]{FEF2F0}74.33 & \cellcolor[HTML]{FEF2F0}\textbf{75.17} & \cellcolor[HTML]{FEF2F0}78.00  & \cellcolor[HTML]{FEF2F0}61.39          & \cellcolor[HTML]{FEF2F0}76.44          & \cellcolor[HTML]{FEF2F0}37.80          & \cellcolor[HTML]{FEF2F0}58.88          & \cellcolor[HTML]{FEF2F0}74.36          & \cellcolor[HTML]{FEF2F0}64.80          & \cellcolor[HTML]{FEF2F0}89.22          \\
                                       &                                            & CosSim(CL)                           & 42.41                                  & 61.15                                  & 77.98                                  & 73.00                                  & 58.80                                  & 67.30                                  & 33.78                                  & 65.27                                  & 72.16                                  & 61.32                                  & 84.42                                  \\
                                       &                                            & \cellcolor[HTML]{FEF2F0}+\textsc{Prune\&Comp} & \cellcolor[HTML]{FEF2F0}\textbf{43.17} & \cellcolor[HTML]{FEF2F0}\textbf{65.57} & \cellcolor[HTML]{FEF2F0}\textbf{86.30} & \cellcolor[HTML]{FEF2F0}\textbf{77.00} & \cellcolor[HTML]{FEF2F0}\textbf{60.82} & \cellcolor[HTML]{FEF2F0}\textbf{69.15} & \cellcolor[HTML]{FEF2F0}\textbf{37.03} & \cellcolor[HTML]{FEF2F0}\textbf{67.40} & \cellcolor[HTML]{FEF2F0}\textbf{77.29} & \cellcolor[HTML]{FEF2F0}\textbf{64.86} & \cellcolor[HTML]{FEF2F0}\textbf{89.29} \\
                                       &                                            & Mag+                                 & 42.75                         & 70.03                         & \textbf{77.00}                         & 77.00                         & \textbf{63.86}                         & \textbf{75.90}                         & 36.46                                  & 58.56                                  & 73.26                                  & 63.87                                  & 87.93                                  \\
                                       &                                            & \cellcolor[HTML]{FEF2F0}+\textsc{Prune\&Comp} & \cellcolor[HTML]{FEF2F0}\textbf{45.31} & \cellcolor[HTML]{FEF2F0}\textbf{75.51} & \cellcolor[HTML]{FEF2F0}72.75 & \cellcolor[HTML]{FEF2F0}\textbf{82.00} & \cellcolor[HTML]{FEF2F0}61.82          & \cellcolor[HTML]{FEF2F0}74.27          & \cellcolor[HTML]{FEF2F0}\textbf{39.52} & \cellcolor[HTML]{FEF2F0}\textbf{60.54} & \cellcolor[HTML]{FEF2F0}\textbf{74.73} & \cellcolor[HTML]{FEF2F0}\textbf{65.16} & \cellcolor[HTML]{FEF2F0}\textbf{89.71} \\
                                       &                                            & Taylor+                              & 40.02                                  & 63.38                                  & 55.90                                  & \textbf{73.00}                         & 62.05                                  & 68.99                                  & \textbf{37.89}                         & 59.98                                  & 71.79                                  & 59.22                                  & 81.53                                  \\
                                       &                                            & \cellcolor[HTML]{FEF2F0}+\textsc{Prune\&Comp} & \cellcolor[HTML]{FEF2F0}\textbf{43.09} & \cellcolor[HTML]{FEF2F0}\textbf{66.08} & \cellcolor[HTML]{FEF2F0}\textbf{82.91} & \cellcolor[HTML]{FEF2F0}72.00 & \cellcolor[HTML]{FEF2F0}\textbf{62.57} & \cellcolor[HTML]{FEF2F0}\textbf{69.75} & \cellcolor[HTML]{FEF2F0}37.51          & \cellcolor[HTML]{FEF2F0}\textbf{62.83} & \cellcolor[HTML]{FEF2F0}\textbf{78.39} & \cellcolor[HTML]{FEF2F0}\textbf{63.90} & \cellcolor[HTML]{FEF2F0}\textbf{87.98} \\
                                       &                                            & CosSim(BI)                           & 46.42                         & 73.74                         & 77.40                                  & 80.00                         & \textbf{64.54}                         & 76.82                                  & \textbf{37.89}                         & 62.75                                  & 76.19                                  & 66.19                                  & 91.13                                  \\
                                       & \multirow{-10}{*}{\textbf{5/36 (11.78\%)}} & \cellcolor[HTML]{FEF2F0}+\textsc{Prune\&Comp} & \cellcolor[HTML]{FEF2F0}\textbf{47.27} & \cellcolor[HTML]{FEF2F0}\textbf{74.07} & \cellcolor[HTML]{FEF2F0}\textbf{81.56} & \cellcolor[HTML]{FEF2F0}\textbf{82.00} & \cellcolor[HTML]{FEF2F0}63.54          & \cellcolor[HTML]{FEF2F0}\textbf{77.09} & \cellcolor[HTML]{FEF2F0}37.80          & \cellcolor[HTML]{FEF2F0}\textbf{63.06} & \cellcolor[HTML]{FEF2F0}\textbf{79.12} & \cellcolor[HTML]{FEF2F0}\textbf{67.28} & \cellcolor[HTML]{FEF2F0}\textbf{92.63} \\ \cmidrule{2-14}
                                       &                                            & PPL                                  & 41.13                                  & 68.64                                  & 67.77                                  & 74.00                                  & \textbf{60.44}                         & 74.59                                  & 35.02                                  & 55.64                                  & 69.23                                  & 60.72                                  & 83.59                                  \\
                                       &                                            & \cellcolor[HTML]{FEF2F0}+\textsc{Prune\&Comp} & \cellcolor[HTML]{FEF2F0}\textbf{43.60} & \cellcolor[HTML]{FEF2F0}\textbf{72.26} & \cellcolor[HTML]{FEF2F0}\textbf{68.23} & \cellcolor[HTML]{FEF2F0}\textbf{77.00} & \cellcolor[HTML]{FEF2F0}58.64          & \cellcolor[HTML]{FEF2F0}\textbf{75.03} & \cellcolor[HTML]{FEF2F0}\textbf{36.84} & \cellcolor[HTML]{FEF2F0}\textbf{56.99} & \cellcolor[HTML]{FEF2F0}\textbf{72.16} & \cellcolor[HTML]{FEF2F0}\textbf{62.31} & \cellcolor[HTML]{FEF2F0}\textbf{85.78} \\
                                       &                                            & CosSim(CL)                           & 33.87                                  & 48.57                                  & 73.55                                  & 69.00                                  & 50.97                                  & 61.86                                  & 31.96                                  & 59.98                                  & 67.03                                  & 55.20                                  & 76.00                                  \\
                                       &                                            & \cellcolor[HTML]{FEF2F0}+\textsc{Prune\&Comp} & \cellcolor[HTML]{FEF2F0}\textbf{35.67} & \cellcolor[HTML]{FEF2F0}\textbf{51.52} & \cellcolor[HTML]{FEF2F0}\textbf{84.59} & \cellcolor[HTML]{FEF2F0}\textbf{72.00} & \cellcolor[HTML]{FEF2F0}\textbf{54.56} & \cellcolor[HTML]{FEF2F0}\textbf{64.53} & \cellcolor[HTML]{FEF2F0}\textbf{34.45} & \cellcolor[HTML]{FEF2F0}\textbf{62.59} & \cellcolor[HTML]{FEF2F0}\textbf{74.36} & \cellcolor[HTML]{FEF2F0}\textbf{59.36} & \cellcolor[HTML]{FEF2F0}\textbf{81.73} \\
                                       &                                            & Mag+                                 & \textbf{40.10}                         & \textbf{68.27}                         & 55.60                         & 71.00                         & \textbf{58.9}                          & \textbf{73.12}                         & 33.59                                  & 56.75                                  & 69.23                                  & 58.51                                  & 80.56                                  \\
                                       &                                            & \cellcolor[HTML]{FEF2F0}+\textsc{Prune\&Comp} & \cellcolor[HTML]{FEF2F0}37.54 & \cellcolor[HTML]{FEF2F0}63.55 & \cellcolor[HTML]{FEF2F0}\textbf{76.33} & \cellcolor[HTML]{FEF2F0}\textbf{75.00} & \cellcolor[HTML]{FEF2F0}55.80          & \cellcolor[HTML]{FEF2F0}71.22          & \cellcolor[HTML]{FEF2F0}\textbf{35.89} & \cellcolor[HTML]{FEF2F0}\textbf{59.04} & \cellcolor[HTML]{FEF2F0}\textbf{71.06} & \cellcolor[HTML]{FEF2F0}\textbf{60.60} & \cellcolor[HTML]{FEF2F0}\textbf{83.44} \\
                                       &                                            & Taylor+                              & 35.75                                  & 49.62                                  & 45.17                                  & \textbf{67.00}                         & 53.29                                  & 64.53                                  & 34.64                                  & 57.54                                  & 64.47                                  & 52.45                                  & 72.20                                  \\
                                       &                                            & \cellcolor[HTML]{FEF2F0}+\textsc{Prune\&Comp} & \cellcolor[HTML]{FEF2F0}\textbf{36.60} & \cellcolor[HTML]{FEF2F0}\textbf{54.50} & \cellcolor[HTML]{FEF2F0}\textbf{85.50} & \cellcolor[HTML]{FEF2F0}\textbf{67.00} & \cellcolor[HTML]{FEF2F0}\textbf{56.31} & \cellcolor[HTML]{FEF2F0}\textbf{65.78} & \cellcolor[HTML]{FEF2F0}\textbf{35.50} & \cellcolor[HTML]{FEF2F0}\textbf{62.51} & \cellcolor[HTML]{FEF2F0}\textbf{72.53} & \cellcolor[HTML]{FEF2F0}\textbf{59.58} & \cellcolor[HTML]{FEF2F0}\textbf{82.03} \\
                                       &                                            & CosSim(BI)                           & \textbf{42.49}                         & 69.07                                  & 68.35                                  & \textbf{80.00}                         & \textbf{61.48}                         & \textbf{75.24}                         & \textbf{35.50}                          & 54.38                                  & 67.77                                  & 61.59                                  & 84.79                                  \\
\multirow{-21}{*}{\rotatebox[origin=c]{90}{Qwen3-8B}}   & \multirow{-10}{*}{\textbf{7/36 (16.49\%)}} & \cellcolor[HTML]{FEF2F0}+\textsc{Prune\&Comp} & \cellcolor[HTML]{FEF2F0}\textbf{40.70} & \cellcolor[HTML]{FEF2F0}\textbf{69.44} & \cellcolor[HTML]{FEF2F0}\textbf{74.07} & \cellcolor[HTML]{FEF2F0}76.00 & \cellcolor[HTML]{FEF2F0}58.86          & \cellcolor[HTML]{FEF2F0}74.70          & \cellcolor[HTML]{FEF2F0}34.93          & \cellcolor[HTML]{FEF2F0}\textbf{57.38} & \cellcolor[HTML]{FEF2F0}\textbf{68.86} & \cellcolor[HTML]{FEF2F0}\textbf{61.66} & \cellcolor[HTML]{FEF2F0}\textbf{84.89} \\ \bottomrule
\end{tabular}%
}
\end{table*}

\paragraph{Magnitude Gap Estimation.}
Once the layer importance is determined by a certain pruning metric, the least-important layer is dropped as Equation \ref{eq:prune}. As motivated earlier, removing any layer introduces a magnitude gap that is independent of the pruning metric.
To compensate for this gap, we estimate the magnitude gap on a small calibration set before pruning. The goal is to obtain an optimal magnitude compensation scalar factor $\alpha$. Assuming that layer $\ell$ is to be removed, $\alpha$ is defined as:s

\begin{equation}
\alpha=\mathbb{E}_{({X}^{(\ell)},{X}^{(\ell+1)})\in \mathcal{D}}\frac{1}{C}\sum_{k=1}^C\frac{\|{X}_{:,k}^{(\ell+1)}\|_1}{\|{X}_{:,k}^{(\ell)}\|_1},
\end{equation}

\noindent where ${X}^{(\ell)}$ and ${X}^{(\ell+1)}$ are the hidden-states at the input and output of layer $\ell$, respectively, collected over the calibration samples.
After $\alpha$ is estimated, we perform layer pruning and scale the input to layer $\ell+1$ with $\alpha$ as compensation.  Formally, the forward propagation with compensation becomes:

\begin{equation}
\label{eq:scale}
{X}^{(\ell+1)}=\alpha{X}^{(\ell)}+f(\alpha{X}^{(\ell)},\theta^{(\ell+1)}).
\end{equation}

\paragraph{Weight Modification.}

Equation \ref{eq:scale} scales hidden states by the scalar $\alpha$ via an element-wise multiplication during inference time, incurring online overhead.  We eliminate this cost by fusing $\alpha$ directly into the weights of model layers that precede the pruned layer. As illustrated in Figure \ref{fig:magcomp}, once $\alpha$ is estimated, it is fused into the model weights in three steps.

\textbf{Step 1: Modifying embedding layer.} The weight of token-embedding layer $W_{embed}$ is updated:

\begin{equation}
W_{embed} \leftarrow \alpha W_{embed}.
\end{equation}

\textbf{Step 2: Modifying MHA output projections.} For every layer with index k$\in$[1, $\ell$-1], the MHA output-projection matrix $W_o$ is updated:

\begin{equation}
W_{o}^{(k)} \leftarrow \alpha W_{o}^{(k)}.
\end{equation}

\textbf{Step 3: Modifying MLP down projections.} Likewise, the down-projection weight $W_{Down}$ of every MLP layer with index k$
\in$[1, $\ell$-1] is scaled:

\begin{equation}
W_{down}^{(k)} \leftarrow \alpha W_{down}^{(k)}.
\end{equation}

We explain the rationale of the weight modification step-by-step. With Step 1, the embedding layer's output hidden states are scaled by $\alpha$. Subsequent Transformer layers consist of a Multi-Head Attention (MHA), a Multi-Layer Perceptron (MLP), and a normalization layer (NORM). The first Transformer layer receives the scaled hidden states from the embedding layer.
The hidden states are then (i) copied as a residual branch, (ii) normalized, (iii) fed to MHA, and (iv) added to the residual. For a normalization layer that is scale-invariant, i.e., NORM(X)=NORM($\alpha$X), we re-introduce the factor $\alpha$ and multiply the output projection weights of MHA by $\alpha$, which effectively scales the MHA output by $\alpha$ before the residual addition. The same logic applies to the MLP block. The scaled MHA output is copied as a residual, normalized, and passed through the MLP. Owing again to normalization's scale invariance, we multiply the down-projection weights of the MLP by $\alpha$, preserving the magnitude. We then apply this procedure to every layer preceding the pruned layer $\ell$.
With these modifications, the model architecture remains unchanged except for the removed layers, incurring no additional online cost at inference time.

\subsection{Iterative Pruning with Magnitude Compensation (\textsc{Prune\&Comp})}

\begin{algorithm}[ht]
    \caption{The proposed \textsc{Prune\&Comp} algorithm.}
    \label{alg:algorithm}
    \footnotesize
    \begin{algorithmic}
        \State $M \gets \text{\textit{original model}}$
        \State $C \gets \text{\textit{calibration dataset}}$
        \State $N \gets \text{\textit{\# blocks of}} M$
        \State $n \gets \text{\textit{\# blocks to remove}}$
        \For{$i = 0$ \textbf{to} $n-1$}
            \State $idx \gets \text{Metric}(M, C)$
            \State $M \gets \text{Prune}(M, idx)$
            \State $M \gets \text{Comp}(M, idx)$
        \EndFor
    \end{algorithmic}
\end{algorithm}

Furthermore, we combine iterative pruning with magnitude compensation. \textsc{Prune\&Comp} iteratively removes layer from an original model $M$ until a target number of $n$ layers has been eliminated, as summarized in Algorithm \ref{alg:algorithm} . In each iteration, the algorithm performs the following three steps:
\begin{itemize}
    \item $Metric(M, C)$: Given model $M$, selects the most redundant layer for removal with a chosen pruning metric over a calibration dataset $C$, and returns its index $idx$. 
    \item $Prune(M, idx)$: Removes the block at index $idx$ from the model $M$ and return the pruned model.
    \item $Comp(M, idx)$: Compensates the model $M$ for the magnitude gap that arises in the hidden states due to layer removal and returns the compensated model.
\end{itemize}
By actively compensating for this gap through training-free magnitude gap estimation and weight modifications, \textsc{Prune\&Comp} ensures that the pruned model maintains a robust internal representation, which in turn enhances the accuracy of subsequent layer importance estimations and ultimately preserves performance. This iterative cycle of pruning and compensation enables the algorithm to effectively reduce model depth while minimizing performance degradation, resulting in a more streamlined and efficient LLM that avoids incurring online inference overhead.

\begin{table*}[t]
\centering
\caption{Performance comparison on MMLU benchmark. 5 layers of LLaMA-3-8B are pruned with CosSim(BI) metric. 
}
\label{tab:table3-mmlu}
\resizebox{1.3\columnwidth}{!}{%

\begin{tabular}{@{}lccccc@{}}
\toprule
\multicolumn{1}{c}{\textbf{Metric}} & \multicolumn{1}{c}{\textbf{STEM}} & \multicolumn{1}{c}{\textbf{Humanities}} & \multicolumn{1}{c}{\textbf{Social Sciences}} & \multicolumn{1}{c}{\textbf{Others}} & \multicolumn{1}{c}{\textbf{Weighted Accuracy}} \\ \midrule
PPL                                 & 28.76                             & 24.36                                   & 27.10                                        & 28.22                               & 26.80                                           \\
\rowcolor[HTML]{FEF2F0} 
+\textsc{Prune\&Comp}                        & \textbf{30.72}                    & \textbf{25.59}                          & \textbf{32.17}                               & \textbf{30.44}                      & \textbf{29.25}                                  \\
CosSim(CL)                          & 53.47                             & 56.08                                   & 74.58                                        & 68.32                               & 62.40                                           \\
\rowcolor[HTML]{FEF2F0} 
+\textsc{Prune\&Comp}                        & \textbf{54.31}                    & \textbf{56.98}                          & \textbf{75.04}                               & \textbf{70.76}                      & \textbf{63.55}                                  \\
Mag+                                & 27.24                             & 23.53                                   & 24.76                                        & \textbf{27.64}                      & 25.54                                           \\
\rowcolor[HTML]{FEF2F0} 
+\textsc{Prune\&Comp}                        & \textbf{28.10}                    & \textbf{24.14}                          & \textbf{31.04}                               & 25.91                               & \textbf{26.91}                                  \\
Taylor+                             & 31.41                             & 37.4                                    & 44.65                                        & 41.39                               & 38.64                                           \\
\rowcolor[HTML]{FEF2F0} 
+\textsc{Prune\&Comp}                        & \textbf{39.36}                    & \textbf{43.97}                          & \textbf{57.26}                               & \textbf{54.90}                      & \textbf{48.42}                                  \\
CosSim(BI)                          & 46.92                             & \textbf{53.92}                          & 65.65                                        & 65.42                               & 57.64                                           \\
\rowcolor[HTML]{FEF2F0} 
+\textsc{Prune\&Comp}                        & \textbf{49.93}                    & 52.09                                   & \textbf{69.45}                               & \textbf{67.46}                      & \textbf{58.98}                                  \\ \bottomrule
\end{tabular}%
}
\end{table*}
\begin{table}[t]
\centering
\caption{Effectiveness of the proposed \textsc{Prune\&Comp}.}
\label{tab:table4-abla}
\resizebox{\columnwidth}{!}{%
\begin{tabular}{@{}lllll@{}}
\toprule
\multicolumn{1}{c}{\textbf{Method}} & \multicolumn{1}{c}{\textbf{WikiText-2}} & \multicolumn{1}{c}{\textbf{C4}} & \multicolumn{1}{c}{\textbf{PTB}} & \multicolumn{1}{c}{\textbf{Average}} \\ \midrule
Naive one-shot pruning   & 57.76                          & 50.13                           & 67.39                            & 58.43                                \\
\rowcolor[HTML]{FEF2F0} 
+IterPrune                & 36.91                          & 32.2                   & 46.00                   & 38.37                       \\
+MagComp                  & 35.38                          & 33.71                           & 43.08                   & 37.39                       \\
\rowcolor[HTML]{FEF2F0} 
+MagComp +IterPrune      & \textbf{28.43}                          & \textbf{24.48}                  & \textbf{28.57}                   & \textbf{27.16}                       \\ \bottomrule
\end{tabular}%
}
\end{table}

\section{Experiments}

\subsection{Experimental Setup and Details}
\paragraph{Models.} We evaluate \textsc{Prune\&Comp} on open-sourced LLMs including LLaMA-2-7B/13B \cite{touvron2023llama2}, LLaMA-3-8B \cite{dubey2024llama3}, Qwen3-8B \cite{qwen3technicalreport}.

\paragraph{Baseline.} We compare with the prevalent layer-wise pruning metrics in one-shot manner, including cosine similarity-based metrics \cite{men2024shortgpt,chen2024compressing,gromov2024unreasonable}, perplexity (PPL)-based metrics \cite{song2024sleb,kim2024shortened}, Taylor+ \cite{kim2024shortened}, and Magnitude+ \cite{kim2024shortened}, as presented in Methods.
To determine the pruned layer and initialize the magnitude compensation factor, 128 sequences with a length of 2048 tokens from the WikiText-2 dataset are randomly selected. All experiments are conducted on a 24GB NVIDIA V100 GPU.

\paragraph{Evaluation.} Three benchmarks are used for evaluation: perplexity (PPL) including WikiText2 \citep{merity2016pointer}, C4 \citep{raffel2020exploring}, and PTB \citep{marcus-etal-1993-building}; Massive Multitask Language Understanding (MMLU) \cite{hendrycks2020measuring}; commonsense question answering (QA) including ARC-Challenge (ARC-c), ARC-Easy (ARC-e) \citep{clark2018think}, BoolQ \citep{clark2019boolq}, HellaSwag (HeSw) \citep{zellers2019hellaswag}, PIQA \citep{bisk2020piqa}, WinoGrande (WG) \citep{ai2:winogrande}, WSC273 (WSC) \citep{levesque2012winograd}, Race-high (Race-h) \citep{lai2017race} and CoPA \citep{sarlin2020superglue}.

\subsection{Results on PPL Benchmark}

Table~\ref{tab:table1-ppl} presents an analysis of the perplexity benchmark for LLaMA-2-7B and LLaMA-3-8B across diverse pruning configurations. We report the average performance on WikiText-2, C4, and PTB datasets.

For the LLaMA-2-7B, the plug-and-play \textsc{Prune\&Comp} consistently improves baselines across all pruning metrics and datasets. For example, when pruning 7 out of 32 layers with CosSim(BI) metric, combining with \textsc{Prune\&Comp} drops the average PPL from 23.57 to 19.88, while with Mag+ metric 
\textsc{Prune\&Comp} improves PPL dramatically from 89.61 to 28.20. This trend of substantial PPL reduction via \textsc{Prune\&Comp} persists at the higher sparsity, demonstrating its effectiveness in mitigating layer pruning-induced performance degradation.
Likewise, LLaMA-3-8B consistently reaps large gains from \textsc{Prune\&Comp}, and combining with every baseline achieves lower perplexity across the metrics. The most striking example is Taylor+ with 5 of 32 layers pruned: its average PPL collapses from an extreme 512.78 to a healthy 16.34, when integrating with \textsc{Prune\&Comp}. The benefit of \textsc{Prune\&Comp} widens when 7 out of 32 layers are pruned. Taylor+ metric, combining with \textsc{Prune\&Comp} plunges from 2840.38 to 25.21, and CosSim(CL) metric, combining with \textsc{Prune\&Comp} likewise tumbles from 2839.30 to 230.73.

\subsection{Results on QA Benchmark}
Table~\ref{tab:table2-qa} presents a comprehensive analysis on question answering (QA) benchmark of LLaMA-3-8B and Qwen3-8B across diverse pruning configurations. 

For LLaMA-3-8B model, the combination of \textsc{Prune\&Comp} consistently enhances the pruned model's performance across all pruning metrics when compared to their baselines, demonstrating its effectiveness. For instance, when pruning 5 out of 32 layers with the Taylor+ metric, the related performance significantly increases from 65.85\% to 90.57\%, leading the baseline with 24.72\%. Similarly, with 7 out of 32 layers pruned, \textsc{Prune\&Comp} raises the related performance from 60.27 to 87.58 using the Taylor+ metric, significantly higher than the baseline.
For Qwen3-8B, a similar positive trend is observed. By integrating \textsc{Prune\&Comp}, the model's relative performance improved across all pruning metrics. For example, when pruning 5 out of 32 layers with CosSim(CL) metric, \textsc{Prune\&Comp} boosts relative performance from 84.42\% to 89.29\%, surpassing the baseline by 4.87\%.

\subsection{Results on MMLU Benchmark}
Further analysis of the massive multitask language understanding (MMLU) benchmark is summarized in Table~\ref{tab:table3-mmlu}. For most sub-tasks, the conventional pruning methods combined with \textsc{Prune\&Comp} yield superior performance, demonstrating their robust capability to improve model performance across diverse knowledge domains.

For example, when pruning 5 out of 32 layers of LLaMA-3-8B with the Taylor+ metric, \textsc{Prune\&Comp} significantly boosts the weighted accuracy from 38.64\% to 48.42\%, leading the baseline by 9.78\%. Likewise, for the CosSim(CL) metric, the performance is improved from 62.40\% to 63.55\%. These results indicate that \textsc{Prune\&Comp} maintains superior performance across multiple pruning metrics.


\subsection{Ablation Studies}
Table ~\ref{tab:table4-abla} investigates the individual contribution of \textsc{Prune\&\\Comp}, including iterative pruning (+IterPrune) and magnitude compression (+MagComp), on the perplexity benchmark.
Both iterative pruning and magnitude compression individually yield significant improvements over naive one-shot pruning across WikiText-2, C4, and PTB datasets. Specifically, with naive one-shot pruning, results in an average perplexity of 58.43, iterative pruning reduces this to 38.37, while magnitude compression achieves an average of 37.39. Crucially, the combination of both achieves the best average perplexity of 27.16, demonstrating a strong synergistic effect where their combined application substantially outperforms individual components.

\section{Conclusion}
We present \textsc{Prune\&Comp}, a training-free plug-and-play layer pruning scheme that mitigates the magnitude gap from layer removal in LLMs via magnitude compensation. By integrating this compensation with iterative pruning, which involves estimating the gap induced by layer pruning and closing it through offline weight rescaling, our method avoids inference overhead while addressing hidden states discrepancy-driven performance degradation.
\textsc{Prune\&Comp} consistently enhances existing layer pruning metrics, validating its effectiveness for practical LLM compression. This work facilitates LLM compression in resource-constrained settings and marks a step toward developing training-free layer pruning schemes.

\nobibliography*


\bigskip

\bibliography{main}



\end{document}